\pdfoutput=1
\PassOptionsToPackage{capitalize}{cleveref}

\documentclass[11pt]{article}

\usepackage{amsfonts,amssymb}
\usepackage[]{acl}
\usepackage[times]{stroop}
\usepackage{inconsolata}
\usepackage{euflag}
\usepackage{listings}

\let\cref\Cref %% VN: i tried to pass [capitalize] to cleveref
\crefname{lstlisting}{listing}{listings}
\Crefname{lstlisting}{Listing}{Listings}
\newcommand{\langpair}[2]{\texttt{#1-#2}}

\renewcommand{\cite}[1]{{\color{red} don't use cite with natbib papers. choose deliberately either citet or citep depending on which is correct}}

\title{The Unreasonable Effectiveness of Random Target Embeddings\\  for Continuous-Output Neural Machine Translation}

\author{Evgeniia Tokarchuk \quad Vlad Niculae \\
   Language Technology Lab \\
   University of Amsterdam \\
  \texttt{\{\href{mailto:e.tokarchuk@uva.nl}{e.tokarchuk}, \href{mailto:v.niculae@uva.nl}{v.niculae}\}@uva.nl}
  }
\begin{document}
\maketitle
\begin{abstract}
Continuous-output neural machine translation (CoNMT) replaces the discrete next-word prediction problem with an embedding prediction.
The semantic structure of the target embedding space (\ie, closeness of related words) is intuitively believed to be crucial.
We challenge this assumption and show that completely random output embeddings can outperform laboriously pre-trained ones,
especially on larger datasets.
Further investigation shows this surprising effect is strongest for rare words, due to the geometry of their embeddings.
We shed further light on this finding by designing a mixed strategy that combines random and pre-trained embeddings,
and that performs best overall.
%for different tokens.
% \renewcommand*{\thefootnote}{*}
% \footnote{Preprint, do not distribute.}
% \renewcommand*{\thefootnote}{\arabic{footnote}}
% \setcounter{footnote}{0}
\end{abstract}

\section{Introduction}
Since text is naturally discrete, \ie, each token in a target sentence is represented by an integer index in the vocabulary, neural machine translation (NMT), as many other language generation tasks, is trained mainly as a discrete-output model with softmax over the full vocabulary followed by the cross-entropy loss. Continuous-output neural machine translation (CoNMT) models, in contrast, are trained to predict the continuous representation based on the distances between vectors. It is an appealing line of study for computational and modeling related reasons~\citep{kumar2018von}, as well as a reliable test bed for exploring the properties of 
% vn: add "language"
continuous language spaces 
that appear in modern deep generative models~\citep{Li-2022-DiffusionLM}.
However, CoNMT introduces its own challenge, namely mapping to and from a continuous space. During training, CoNMT model requires continuous targets, and while decoding, one needs to map back to the discrete text representation.

Text mapping to continuous space is widely explored in NLP and can be done using \textit{embeddings} of tokens, words~\citep{turian-etal-2010-word,mikolov-w2v,mikolov-etal-2018-advances} and sentences~\citep{reimers-gurevych-2019-sentence,feng-etal-2022-language}.
Cosine similarity between word embeddings is well correlated with lexical similarity metrics, motivating the use of cosine distance against pre-trained embeddings
%as the dominant training strategy for CoNMT. 
as an effective training strategy for CoNMT
Nearest neighbor beam decoding would in this case include related words and, unlike discrete cross-entropy, the training strategy does not discourage synonyms.

Previous studies show that the quality of continuous-output models highly depends on the choice of embeddings~\citep{Li-2022-DiffusionLM, tokarchuk-niculae-2022-target, kumar2018von}. In general, in CoNMT the embeddings are pre-trained and fixed: otherwise, making all embeddings equal yields an unwanted global optimum. Obtaining pre-trained word embeddings can be computationally expensive, especially if one needs to train an embeddings model from scratch.

In this work we \textit{randomly} initialize target embeddings for continuous-output models and keep them static during training.
\citet{arora-etal-2020-contextual} applied static random embeddings for text classification model's input; however, to the best of our knowledge, the effect of untrained random target embeddings has not been previously studied in the literature, especially for text-generating tasks such as machine translation.
%Using random embeddings as targets for CoNMT training
%challenges the assumption that semantic relationships between word embeddings are necessary.
%confronts the idea of the semantic similarity importance. 
However, we show that random target embeddings perform close to their pre-trained counterpart, and even surpass them 
on the larger datasets, challenging the assumption that target embeddings must preserve semantic relationships.
%if there is enough data available. 
%\et{here rev2 complained about meaningful semantic similarity cause we use MTTransfer} That means that meaningful semantic similarity is not the only factor contributing to the performance of the continuous-output models.
%\et{replacement start here}
Meaningful structures in target embedding space could help with generalization, but our results suggest that any such benefits are smaller than one might expect, and sensitive to embedding concentration.
%\et{replacement ends here}
We hypothesize and bring experimental evidence that CoNMT performance is negatively impacted when there is too little space around embeddings, \ie, when embeddings are tangled rather than more spread out.
%that distances between embeddings 
%that 
%play an important role for representation disentanglement. 
Our findings on three NMT tasks, namely WMT 2018 English$\rightarrow$Turkish (\langpair{en}{tr}), WMT 2016 English$\rightarrow$Romanian (\langpair{en}{ro}), and WMT 2019 English$\rightarrow$German (\langpair{en}{de}) indicate that random embeddings
are more spread out and perform better on rare words for all language pairs. 
%% VN -- small rephrasing here
Strikingly, on the largest dataset (\langpair{en}{de}), random embeddings show the largest gain over pre-trained ones.
%On the large-scale (\langpair{en}{de}) CoNMT with random target embeddings are even substantially better overall. 
We propose a simple yet efficient combination of random and pre-trained embeddings, and show that it improves model performance 
in most cases considered.
%on both \langpair{en}{tr} and \langpair{ro}{en}
More generally, our findings show that dispersion is an important property of embedding space geometry,
and that integrating semantic information should be done with care.

\section{Continuous-Output NMT}
The machine translation task involves learning to map sequences of input tokens
\(\bm{x} = (x_1, \ldots, x_m)\) to output tokens \(\bm{y}=(y_1, \ldots, y_n)\).
In standard (discrete) NMT,
each step is a multi-class next word prediction task, minimizing:
\begin{equation}\label{eq:discrete}
\begin{gathered}
L_\text{discrete}(y_i=t; \bm{y}_{<i}, \bm{x}) = -\log p(y_i=t \mid \bm{y}_{<i}, \bm{x}) \\ = -\DP{\bm{E}(t)}{\bm{h}} + \log\sum_{t' \in V}\exp \DP{\bm{E}(t')}{\bm{h}},
\end{gathered}
\end{equation}
where \(t\) is a token index, \(V\) is the vocabulary, \(\bm{E} : V \to \bbR^d \)
is an embedding lookup,
and \(\bm{h}\) is a transformer hidden state calculated in terms of  \(\bm{x}\) and the output prefix \(\bm{y}_{<i}\).
The costly log-sum-exp and the penchant for continuous similarity metrics in NLP
motivate a purely-continuous alternative:
\begin{equation}\label{eq:cosineloss}
    L_\text{cos}(y_i=t ; \bm{y}_{<i}, \bm{x})  = 1 - \operatorname{cos}(\bm{E}(t), \bm{h}).
\end{equation}
Continuous NMT models were first studied by
\citet{kumar2018von}, who also propose other probabilistic losses and
later other margin-based objectives \citep{bhat-etal-2019-margin},
with limited gain and at the cost of
additional hyperparameters; we therefore focus on the robust cosine objective. We further justify the choice of cosine over max-margin as an objective function in~\cref{app:max-margin}.

On the other hand, the choice of embeddings \(\bm{E}\) makes a much larger difference,
especially due to the fact that
all previous work keeps this parameter frozen: indeed, if it were trainable,
\cref{eq:cosineloss} would have trivial global optima by setting all
\(\bm{E}(t)\) to the same vector for all \(t\).
With modern transformer architectures, the best performing embeddings overall tend to be the
``oracle'' output embeddings learned by a pre-trained discrete MT system
\citep{tokarchuk-niculae-2022-target}.
We highlight that the cosine loss is invariant to the norms of both the embeddings and of the decoder hidden state,
and therefore we may restrict our modeling problem to the unit sphere.

Optimizing \cref{eq:discrete} pushes the model \(\bm{h}\) away from all tokens different from the ``gold'' token,
even if some other tokens (\eg, synonyms) could otherwise be a good fit. \cref{eq:cosineloss} has no such effect,
leading to a promise of more diverse generations.
%\et{next paragraph is revised}
An appealing intuition is that synonyms and related words being nearby in embedding space contributes to the performance of CoNMT and enables such diversity. However in practice, greedy nearest-neighbor lookup is applied, and beam search decoding is not well-studied in the context of CoNMT. Therefore, in this work, we dwell more into the beam search performance for CoNMT, and compare pre-trained and completely random embeddings.
%Therefore, in this work, we challenge this conventional wisdom by considering completely random embeddings.

%is usually done by greedy nearest-neighbor lookup rather than beam search.
%\et{here reviewer pointed: The way you describe the context here slightly contradicts your findings regarding beam search. I would suggest rephrasing this part.}However, this intuition is not consistent with practice.

\section{Random Embeddings Generation}
We consider two different distributions from which to sample the $|V|$ random embeddings.
\paragraph{Spherical uniform.}
We draw embeddings uniformly from the surface of the sphere: \(\mbE(y_i) \sim \operatorname{Unif}(\bbS_{d-1})\).
Since standard normal vectors are distributed with rotational symmetry around the origin,
uniform samples on the sphere can be obtained by normalizing standard normal random vectors:
\[\mbE(y_i) = \mbu_i / \|\mbu_i\|; \quad \mbu_i \sim \operatorname{Normal}(\bm{0}, \bm{I}_d).\]
The same argument works if the normal distribution has spherical covariance \(\sigma \bm{I}_d\)
for any \(\sigma\), and thus, since the cosine loss is norm-invariant,
uniform initialization is exactly equivalent to the standard initialization of transformer embeddings.
%(For the cosine loss, this is equivalent to initializing embeddings from an iid Gaussian \(\mathcal{N}(0, \sigma^2 I)\)
%and keeping them frozen, as the cosine loss is norm-invariant.)

\paragraph{Hypercube.}
The corners of the hypercube \(\{-1, 1\}^d\) all have norm \(\sqrt{d}\) and thus form a discrete subset of a hypersphere.
This motivates us to consider drawing embeddings from a scaled Rademacher distribution:
\begin{align*}
\mbE(y_i) = \mbr_i /\sqrt{d};
\quad
\mbr_i \sim \operatorname{Rademacher}(d).
\end{align*}
Each coordinate of \(\mathbf{r}_i\) has 50\% probability of being \(+1\) and 50\%
of being \(-1\).
With this strategy, any two distinct embeddings have cosine distance at least \(2/d\).
Moreover, hypercubic embeddings can be stored as bit patterns and potentially allow for faster loss calculation with dedicated low-level implementations
which we do not explore here.

\begin{table*}[ht]
    \centering
    \resizebox{0.85\linewidth}{!}{%
    \begin{tabular}{l c c c c c c}
    \toprule
                                  &  \multicolumn{2}{c}{\langpair{en}{tr}}  & \multicolumn{2}{c}{\langpair{ro}{en}}      &\multicolumn{2}{c}{\langpair{en}{de}} \\
\textbf{embeddings}   & BLEU $\uparrow$ & BERTSc. $\uparrow$ &  BLEU $\uparrow$ & BERTSc. $\uparrow$   & BLEU $\uparrow$ & BERTSc. $\uparrow$\\ \midrule
discrete model        & 12.3 & 70.4                          & 31.7 & 64.1                             & 33.1 & 69.0\\ \midrule
pre-trained (beam=1)   & 10.1 & 67.1                          & 29.0 & 58.5                            & 31.3 & 66.2 \\
pre-trained            & \textbf{10.4} & 67.4                          & 29.0 & 58.0                             & 29.2 & 62.6 \\
random uniform        & 8.9 & 65.1                           & 28.8 & 58.8                             & 31.8 & \textbf{67.2}\\
random cube           & 8.7 & 64.6                           & 28.7 & 58.8                             & 31.4 & 66.9\\
%MTtransfer+random (beam=1)    & 10.4 &  67.5 & \textbf{29.6} & \textbf{60.5} & 30.9 & 66.0\\
combined   & \textbf{10.4} & \textbf{68.3} & \textbf{29.5} & \textbf{60.4} & \textbf{32.0} & 66.8 \\ \bottomrule
% combined  $\alpha=0.7$   & 10.4 & 68.2 & 29.3 & 60.1 &  31.9 & 66.4 \\        \bottomrule
    \end{tabular}%
    }

    \caption{BLEU and BertScore on \langpair{ro}{en} \texttt{newstest16}, \langpair{en}{tr} \texttt{newstest2017} and \texttt{newstest2016} \langpair{en}{de}. We use a beam of 5 if not stated otherwise. In bold, we show the highest score among the continuous models in each column.}
    \label{tab:main results}
\end{table*}

\section{Experimental Setup and Data}
We train CoNMT systems with pre-trained target embeddings as well as randomly-generated target embeddings. 
The \textbf{pre-trained embeddings} we use are extracted from a discrete NMT system trained on the same training data, following the setup of \citet{tokarchuk-niculae-2022-target}, who found this strategy to outperform other subword-level pre-trained embeddings for CoNMT.
% \vn{Do we need to give any other info here about training? or is it exactly as in the wkshp paper}

%and against pre-trained embeddings. Pre-trained embeddings are extracted from well-trained discrete NMT system as in a same way as in~\citet{tokarchuk-niculae-2022-target}.

Results are reported on three WMT translation tasks:\footnote{https://www2.statmt.org/} WMT 2016 Romanian$\rightarrow$English (\langpair{ro}{en}), WMT 2018 English$\rightarrow$Turkish (\langpair{en}{tr}) and WMT 2019 English$\rightarrow$German (\langpair{en}{de}), the latter including back-translated data. Note that for \langpair{en}{tr} we use only WMT 2018 training data with 207k training sentences 
%to represent 
in order to investigae
a challenging lower-resource and morphology-rich scenario. Data statistics are collected in \cref{app:data-stats}.

For subword tokenization we used the same SentencePiece ~\citep{kudo-richardson-2018-sentencepiece} model for all language pairs, specifically the one used in the mBart
multilingual model
~\citep{liu-etal-2020-multilingual-denoising}.
This choice allows for unified preprocessing for all languages we cover. We validate that token-based models performs generally better than word-level models (\cref{app:word-level-emb}), even though subwords introduce an additional challenge of predicting subword continuation (\cref{app:subword-emb}).

We used the \texttt{fairseq}~\citep{ott-etal-2019-fairseq} framework for training our models. Baseline discrete models are trained with cross-entropy loss, label smoothing equal to 0.1 and effective batch size 65.5K tokens. Both discrete and continuous models are trained with learning rate $5\cdot10^{-4}$, 10k warm-up steps for \langpair{ro}{en} and \langpair{en}{de}, and 4k for the smaller \langpair{en}{tr} dataset. All continuous models are trained with the cosine distance objective in \cref{eq:cosineloss}. 
We provide all training details in~\cref{app:training-details}.
%Full details of training 
%Detailed description of training setup and parameters can be found in~\cref{app:training-details}.

We measure translation accuracy using SacreBLEU
\footnote{nrefs:1|case:mixed|eff:no|tok:13a|smooth:exp|version:2.3.1}
\citep{papineni-etal-2002-bleu,post-2018-call}
and BertScore\footnote{implementation by https://github.com/Tiiiger/bert\_score} \citep{Zhang2020BERTScore}. Note that BertScore is scaled differently for each language,
so the scores cannot be compared across languages.

\begin{figure}[t]
    \centering
    \includegraphics[width=0.42\textwidth]{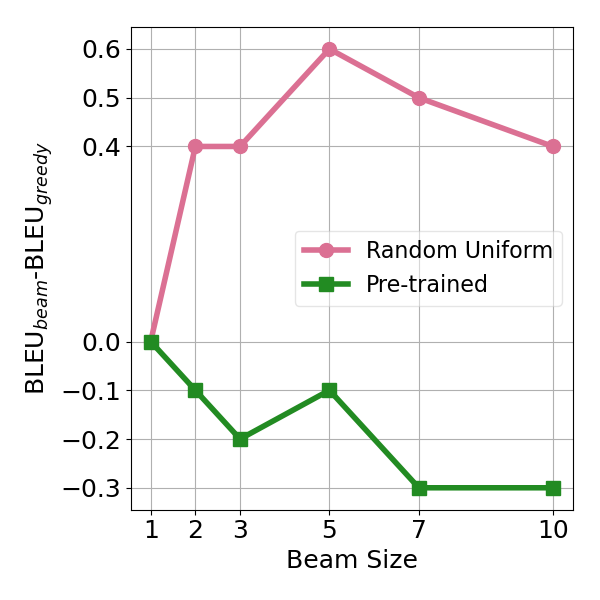}
    \caption{{BLEU}$_\text{beam}-${BLEU}$_\text{greedy}$ scores for the \langpair{ro}{en} \texttt{newsdev2016} for continuous output models with uniform random and pre-trained embeddings. Greedy (beam size 1) BLEU scores are 30.0 for pre-trained, and 28.6 for random embeddings.}
    \label{fig:beam-bleu}
\end{figure}

\section{Results and Discussion}
\label{sec:results}
\paragraph{Scores.}
Per \cref{tab:main results}, we find that random uniform embeddings outperform the pre-trained baseline for \langpair{en}{de},
match it closely for \langpair{ro}{en}, and only underperform in the low-resource case for \langpair{en}{tr}.
We find that hypercube embeddings consistently perform no better than uniform embeddings; however, it is possible that their computational advantages
can make up for this in some applications.

\paragraph{Beam search.}
%We note from \cref{tab:main results} that, except for \langpair{en}{tr}, the MTtransfer model performs
%worse when using $\text{beam size}>1$. But on random uniform embeddings we observe the opposite: beam search is at least as good as greedy search, usually improving.
Preliminary experiments with CoNMT models indicate little gain or even degradation from beam search,
which is why we also report results with greedy decoding for pre-trained in \cref{tab:main results}.
Further investigation in \cref{fig:beam-bleu} shows that
%despite the expectation
%despite the meaningful distances in embedding space,
the 
%pre-trained model 
\langpair{ro}{en} model with pre-trained embeddings
degrades consistently, performing best in the greedy case, while the random embedding model
benefits noticeably from a larger beam, in spite of neighboring words being random and not related.
We discuss the details of the beam search in~\cref{app:beam-search}.
\begin{figure*}[ht]
            \centering
            \includegraphics[width=0.98\textwidth]{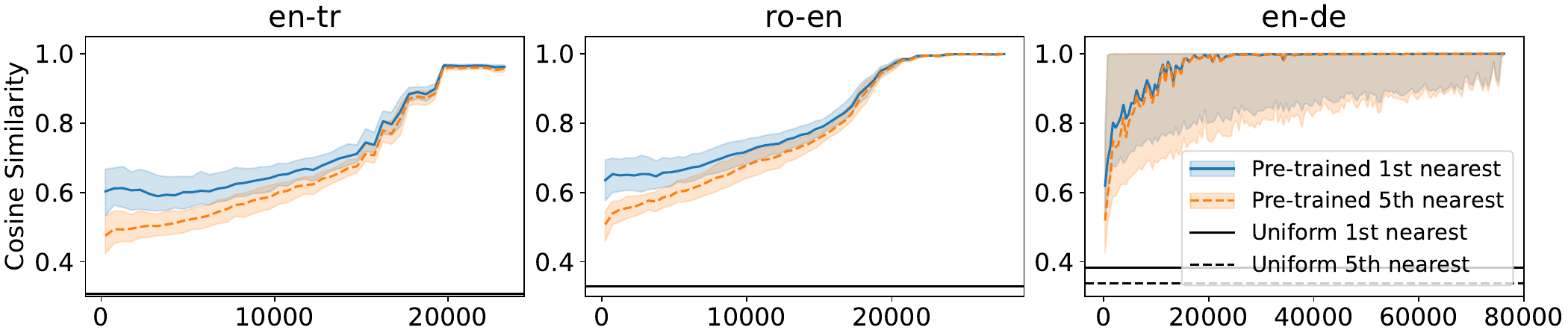}\\
            \includegraphics[width=0.98\textwidth]{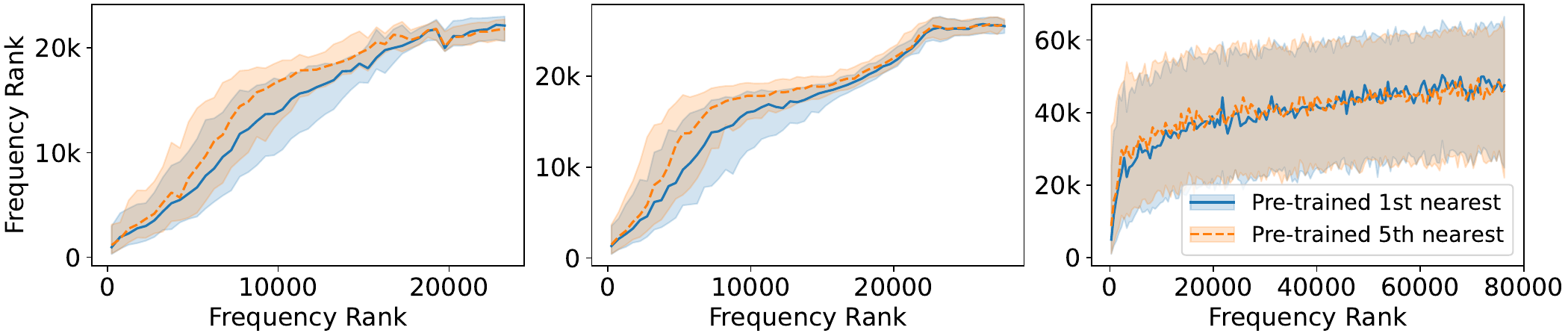}
        \caption{\label{fig:sims-emb}
Pre-trained embeddings demonstrate strong correlation between the frequency rank of each token and (top) the cosine similarity, and (bottom) the frequency rank of its nearby neighbors. Most rare words are identified with their nearest neighbor, which is also a rare word.
Bin size 500; shaded area denotes 50\% of values in each bin.}
\end{figure*}
\paragraph{Frequency.}
\begin{figure}[ht]
    \centering\hspace{-.7cm}
    \includegraphics[width=0.521\textwidth]{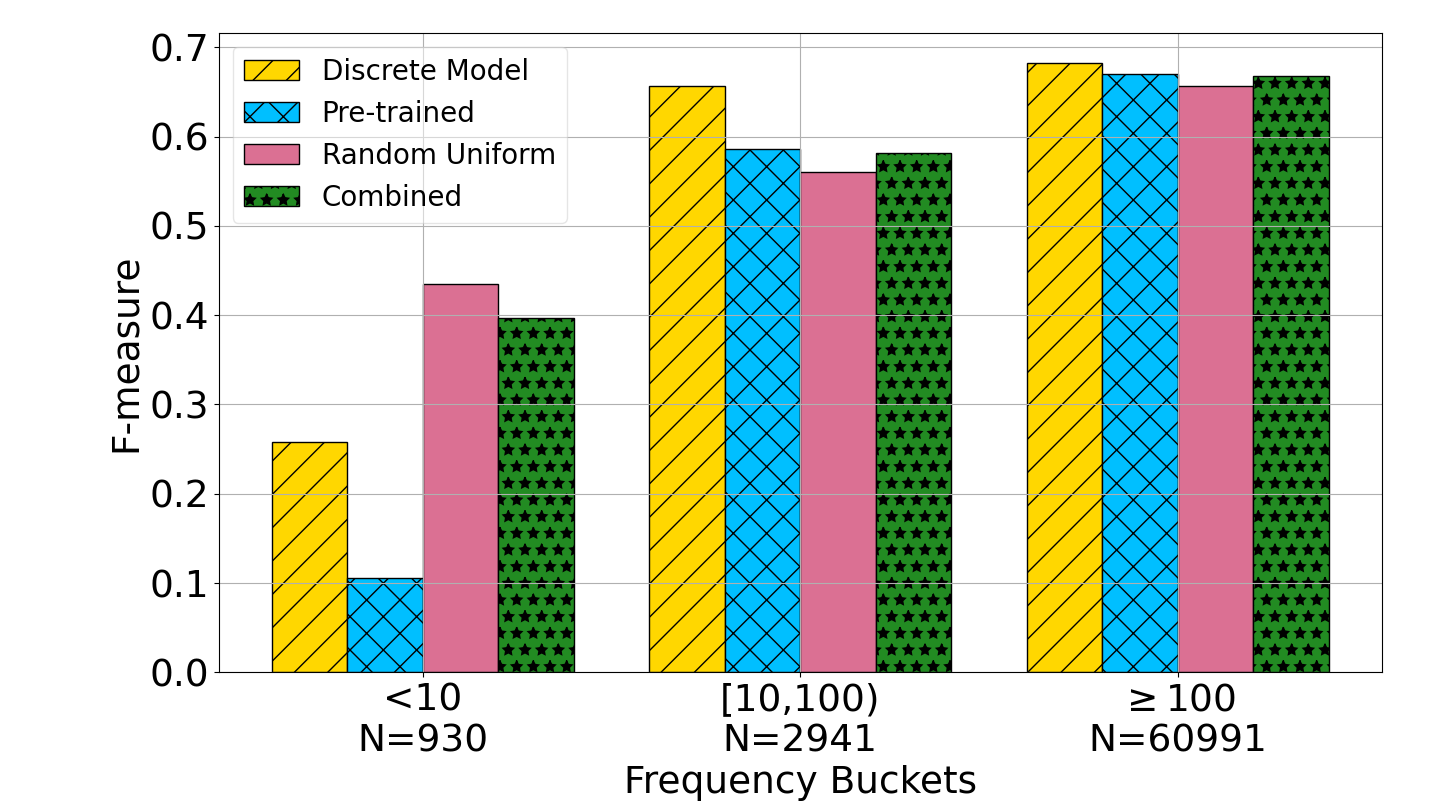}
    \caption{%
Token-level \(F_1\) test score grouped into three bins defined by training set frequency.
The \(x\) label shows frequency boundaries and token counts per bucket. 
%F-measure on tokens level by frequency bucket. Frequency is defined by the occurrence of  \texttt{newstest2016} tokens in \langpair{ro}{en} training set. We compare the best models for each category.
}
    \label{fig:score-by-freq}
\end{figure}
We perform a token-level evaluation using
%A look at \(F_1\) scores binned by training-set frequency
%using
\texttt{compare-mt}~\citep{DBLP:journals/corr/abs-1903-07926},
computing the \(F_1\) score of matching a gold token (at its gold position),
aggregated over bins defined by the token's frequency in the training data.
The result in \cref{fig:sims-emb}
reveals that random embeddings allow much better classification of rare tokens than even the discrete reference model.
To understand this effect, we study the geometry of the pre-trained embedding spaces in relation to frequency in \cref{fig:sims-emb}.
The top row shows the relationship between the frequency
\emph{rank} (higher means rarer) and the similarity to its nearest-- and fifth-nearest-- neighbors.
For all three language pairs we observe that most rare words become identical to their nearest neighbor. In contrast, for random embeddings this metric
does not depend on rank and is always around 0.4.
The bottom row of \cref{fig:score-by-freq} shows that the nearest neighbors of rare words tend also to be comparably rare.
This geometry clarifies in part the surprising performance of random embeddings
on rare tokens.

\paragraph{Combined embeddings.}
Our finding motivates
combining pre-trained and random embeddings:
\begin{equation*}\label{eq:ptsumrand}
%    \mathbf{E}(y_i) = \frac{1}{Z_{cmb}} \Bigl( \alpha \mathbf{E}_{MT}(y_i) + (1-\alpha)\mathbf{E}_{rand}(y_i) \Bigr),
\mbE_\text{cmb}(y_i) = \frac
{\alpha \mbE_\text{pre}(y_i) + (1-\alpha)\mbE_\text{rand}(y_i)}
{\|\alpha \mbE_\text{pre}(y_i) + (1-\alpha)\mbE_\text{rand}(y_i)\|}.
\end{equation*}
%where $Z_{cmb}$ is equal to the norm of convex combination.
%To avoid putting more emphasis to the noise and keep pre-trained distances,
To emphasize pre-trained distances more than the noise,
we choose $\alpha=0.9$ for all language pairs. This simple approach leads to overall improved performance, on almost all metrics and language pairs as shown in ~\cref{tab:main results}. Furthermore, ~\cref{fig:score-by-freq} confirms that combined embeddings preserve the performance of pre-trained embeddings on frequent tokens and increase \(F_1\) score on rare tokens.
We further study the impact $\alpha$ on \langpair{ro}{en} in ~\cref{app:combined-emb} and observe that for all considered $\alpha \in [0.5,0.9]$, the combination outperforms random and pre-trained embeddings along both metrics;
the specific value of \(\alpha\) in this range has only negligible impact.

%ones on both metrics.
%In~\cref{tab:main results} we report these results under the name \texttt{MTtransfer+random}.
%replacing the embeddings of rare words with random embeddings while keeping pre-trained embeddings for more frequent words. We choose the threshold based on token's rank, precise values for each language pairs are reported in~\cref{app:combined-emb}.

% However, on \langpair{en}{de} results are inconsistent for provided metrics.

%On \langpair{en}{de} performance degrades, perhaps due to a suboptimal choice of threshold.
\section{Additional Related Work}
\paragraph{CoNMT losses.}
%We rely in our investigation on the work of~\citet{tokarchuk-niculae-2022-target}.
Earlier work in CoNMT suggests loss functions other than cosine, based on
modified Langevin (a.k.a.~von Mises-Fisher) log-likelihood,
or based on max-margin constructions, to perform better
\citep{kumar2018von,bhat-etal-2019-margin}.
Nevertheless, in preliminary experiments, we find that when using more modern
architectures and datasets, such objectives do not outperform the cosine loss.
%Thus we focused on the simpler and more numerically stable objective function in our work.
The cosine loss is an instance of Langevin log-likelihood
with spread $\kappa=1$ (\cref{app:beam-search}), allowing for a
theoretically-grounded beam search over sequence likelihood, whereas for
max-margin losses it is not clear how to derive a principled beam search.
Nevertheless, we provide a small set of additional experiments confirming that
max-margin losses underperform cosine while showing similar effects in
\cref{app:max-margin}.
 %at token level, cosine and vMF only differ by a weighting
%induced by $\kappa$ and additive constant.
%While ~\citet{kumar2018von} showed
%that using von Mises-Fisher loss (vMF) loss is beneficial for the final
%performance, our preliminary investigation showed that
%~\cref{app:beam-search} the beam
%search is theoretically grounded as a comparison of sequence likelihoods. Beam
%search indeed gets tricky for non-probabilistic losses like max-margin, which
%we therefore deemed out of scope for this work; we nevertheless are interested
%in studying margin-based losses in future work. We believe vMF to exhibit the
%same effects while using random embeddings. We leave max-margin losses out of
%scope was their lack of a probabilistic interpretation, preventing correct
%application of beam search.

\paragraph{Retrieval-augmented NMT.}
%\et{here: changed the wording/motivation kind of}  % looks good to me
Similarly to CoNMT, $k$-NN MT~\citep{khandelwal2021nearest,yogatama-etal-2021-adaptive,stap-monz-2023-multilingual} relies on the distance-based retrieval from datastore in decoding time, with cosine similarity and Euclidean distance as a popular choice of the similarity measure. Even though creation of a datastore and extracting target embeddings are two distinct processes, they both share similar traits and rely on discrete transformer MT system as a source of representations. \citet{li_better_2022} argue that quality of $k$-NN MT directly depends on the quality of retrieved neighbors contexts from the datastore, and show that $k$-NN MT exhibits a related issue with high similarity between unrelated keys. 
Our findings suggests that randomization could provide paths toward improved performance in $k$-NN MT. 
%\vn{Directly applying our finding to randomize the datastore is not feasible, as it would require fine-tuning and impose distance even between identical contexts. However, other}
% \vn{Suggested phrasing for here, please edit} The construction of our target embeddings,
% motivated by a transfer learning angle \citep{tokarchuk-niculae-2022-target}, is
% related to---but different from---the construction of the data store in
% retrieval-augmented NMT
% \citet{khandelwal2021nearest,yogatama-etal-2021-adaptive,stap-monz-2023-multilingual}.
% In this line of work, \citet{li_better_2022} identified a related issue with high similarity between
% unrelated keys. We speculate that frequency and randomness might play similar
% roles in that setting.
\paragraph{Unargmaxability.}
\citet{grivas-etal-2022-low} point out that standard (discrete) language models can
have ``unargmaxable'' vocabulary items. When using directional modelling (on the
unit sphere), unargmaxability is mitigated and only occurs for identical
embeddings; however, embeddings that are too close to their neighbors can have
very small Voronoi sets, leading to the phenomenon we identify in this work,
which is problematic in practice for CoNMT. Random perturbations to embeddings
might effectively mitigate unargmaxability in discrete models as well.

\paragraph{Hubness.}
Hubness~\citep{Dinu2014ImprovingZL,lazaridou-etal-2015-hubness,huang-etal-2019-hubless} is a phenomenon that impacts nearest-neighbor retrieval as well, characterized by the presence of a few data points (hubs) that are close to many other data points despite their semantic dissimilarity. 
The phenomenon we observe is related but different: many rare words are embedded very close to another rare word, but
not necessarily close to all others overall. Therefore, methods for reducing hubness would not necessarily prevent this situation.

\section{Conclusion}
Our experimental results show that randomly initialized target embeddings can achieve similar performance as pre-trained ones and even surpass them when a sufficiently large amount of data is available. The gap is most pronounced on very rare tokens. We also found that $\text{beam size}>1$  does not harm the performance of CoNMT with random target embeddings (compared to pre-trained target embeddings).
We suggest combining random and pre-trained embeddings in attempt to maintain high accuracy on frequent tokens as well as rare tokens. This simple approach proved to be effective for \langpair{en}{tr} and \langpair{ro}{en} in terms of overall performance.
However, more refined ways to combine random embeddings with semantically meaningful anchors may lead to more reliable improvements, and ideally
hold the potential to remove the reliance on a pre-trained model entirely.
Finding the best ways to achieve this potential is an important avenue of future work for CoNMT and for continuous modeling of language repesentations more broadly.

\section*{Limitations}
\paragraph{Generalization.} Our experimental results show that semantic similarity of the targets embeddings does not play a major role for continuous-output NMT. However, this may not necessarily hold for other text generation tasks like summarization or language modeling. To claim that random target embeddings can be sucessfuly used for any text generation task yet has to be proved. In the future, we will conduct additional experiments on other text generation tasks, such as summarization and language modeling. \par

\paragraph{Dataset Size.} ~\citet{arora-etal-2020-contextual} argue that random embeddings can achieve comparable performance when the dataset size is big enough. In our work we report results on three language pairs with vast range of training samples.The gap between pre-trained and random embeddings is much higher for \langpair{en}{tr} with 207K training samples than for \langpair{ro}{en} and \langpair{en}{de} with 612K and 9.1M training samples. Moreover, on \langpair{en}{de} random embeddings even outperform pre-trained ones.
That hints that random embeddings indeed work only if there is sufficiently large amount of data available. \par

\paragraph{Static Embeddings.} The formulation of the loss we use in our work, specifically cosine distance, leads to representation collapse when tuning target embeddings jointly with the model, That is why in our work the target embeddings are kept unchanged during training. ~\citet{Li-2022-DiffusionLM} show that it is possible to design a loss that allows for joint training. However, we believe that fine-tuning of random embeddings is orthogonal to our study. \par

\paragraph{Comparison with External Embeddings.}
In the scope of this work, we compared only embeddings extracted from the discrete NMT model (pre-trained) and randomly generated embeddings. However, we do not compare random embeddings with external models like mBart~\citep{liu-etal-2020-multilingual-denoising} or fasttext~\citep{bojanowski-etal-2017-enriching}. That is intentional since ~\citet{tokarchuk-niculae-2022-target} showed that pre-trained embeddings extracted from discrete NMT system perform the best compared to the external models, and our goal was to compare to the best-performing baseline. \par

\paragraph{Loss Function.} All our results are tied to the choice of the target objective function, precisely cosine similarity. We chose cosine similarity to align our work with previous studies on CoNMT~\citep{kumar2018von, tokarchuk-niculae-2022-target}. Although our preliminary experiments with Langevin-based as well as with margin-based losses suggested worse performance than cosine for CoNMT, 
other less-studied objectives, \eg, based on geodesic distances, or on expectations of a discrete loss
\citep{Scott2021vonML}, left outside of our scope, may lead to further improvement.
%will help to verify if our findings hold true.
%\et{updated text a bit}
%We implicitly assumed that our embeddings lie on the sphere and have the norm equal to 1. In the future, we would like to experiment with other geometrical spaces and verify if our findings are still valid.
% \vn{Update this to avoid repetition with related work, maybe remove?}

\section*{Risks}
NMT as a technology is subject to dual-use concerns. We also want to stress that it is possible that random embedding models make different kinds of mistakes compared to other models, and they should be studied and treated with caution before deployment. CoNMT models are generally at an earlier stage of development
and do not seem likely to replace the well-studied discrete models in deployed application in the very near future.

\section*{Acknowledgements}
We thank all the members of the UvA Language
Technology Lab for their valuable feedback on
our work. Special thanks to  Kata Naszádi, Sergey Troshin, Caio F.~Corro, Andreas Grivas, and Wilker Aziz for their time and insightful comments. We also thank SURF (www.surf.nl) for the support in using the National Supercomputer Snellius.
This work was partly supported by
the Dutch Research Council (NWO) via VI.Veni.212.228 and
the European Union's Horizon Europe research and innovation programme
via UTTER 101070631.
\hfill{\huge\euflag[-.5\baselineskip]}

\bibliography{anthology,custom}
\appendix
\section{Data Statistics}
~\cref{tab:data-stats} contains data statistics for datasets used in our experiments.
\label{app:data-stats}
\begin{table*}[ht]
    \centering
    \resizebox{\linewidth}{!}{%
    \begin{tabular}{l|c|c|c|c|c|c|c|c|c|c|c}
    \toprule
            \multirow{2}{*}{} & \multicolumn{3}{c|}{WMT \langpair{ro}{en}} &  \multicolumn{4}{|c}{WMT \langpair{en}{tr}} &   \multicolumn{4}{|c}{WMT \langpair{en}{de}} \\

            & train & dev16 & test16 &  train & dev17 & test17 & test18 & train & valid & test16 & test18\\
            \midrule
            sentences & 612K& 2K & 2K & 207K& 1K & 3K & 3K & 9.1M & 2.2K & 3K & 3K\\ \midrule
            SPM vocabulary (tgt) & \multicolumn{3}{c|}{27.5K} & \multicolumn{4}{c|}{23.3K} &\multicolumn{4}{c}{76K}\\\cline{2-12}
           SPM \% oov (tgt)  & 0.0 & 0.38  & 0.31 & 0.0&0.45 &0.53 & 0.55 &0.0&0.0&0.0&0.0\\

          \bottomrule

    \end{tabular}%
    }
    \caption{Datasets Statistics}
    \label{tab:data-stats}
\end{table*}

\section{Models' Training Parameters}
\label{app:training-details}
We report \texttt{fairseq} yaml config in ~\cref{list:yaml-config}.
%Parameters unique for the specific language is highlighted by \{parameter\}\_\{language\_pair(s)\}.
Language-pair--specific parameters are highlighted with a comment.
Continuous transformer uses base Transformer architecture with 6 layers of encoder and decoder~\citep{Vaswani-trafo}. Total number of training parameters is the following: \langpair{ro}{en} discrete is 42M and \langpair{ro}{en} continuous 74M; \langpair{en}{tr} discrete is 40M and \langpair{en}{tr} continuous 73M; \langpair{en}{de} discrete is 132M and \langpair{en}{de} continuous 123M.

\definecolor{mygreen}{rgb}{0,0.6,0}
\definecolor{mygray}{rgb}{0.95,0.95,0.95}
\lstdefinestyle{yaml}{
     basicstyle=\color{mygreen}\footnotesize,
     tabsize=2, 
     rulecolor=\color{black},
     string=[s]{'}{'},
     stringstyle=\color{mygreen},
     comment=[l]{:},
     commentstyle=\color{black},
     morecomment=[l][\color{blue}]{\#},
     linewidth={0.5\textwidth},
     frame=tb,
     framerule=1pt
 }

\begin{lstlisting}[language=Python, 
  caption={Training yaml config for CoNMT},
  label={list:yaml-config},
  captionpos=t,
  style=yaml,
  ]
task:
  _name: translation
  data: language_specific_data
criterion:
  _name: cosine_ar_criterion
model:
  _name: continuous_transformer
  decoder:
    output_dim: 128
    learned_pos: true
  encoder:
    learned_pos: true
  dropout: {0.1, 0.3}
    # 0.3 for ro-en and en-tr 
    # 0.1 for en-de
  target_embed_path: $PATH
  no_decoder_final_norm: false
optimizer:
  _name: adam
  adam_betas: (0.9,0.98)
lr_scheduler:
  _name: inverse_sqrt
  warmup_updates:{10000,4000}
  	 # 10000 for ro-en and en-de
  	 # 4000 for en-tr
  warmup_init_lr: 1e-07
dataset:
  validate_after_updates: 10000
  max_tokens: 4096
  validate_interval_updates: 2000
optimization:
  lr: [0.0005]
  update_freq: [16]
  max_update: 50000
  stop_min_lr: 1e-09
checkpoint:
  no_epoch_checkpoints: true
  best_checkpoint_metric: bleu
  maximize_best_checkpoint_metric: true
\end{lstlisting}

We train our models using shared GPU cluster, which is equipped with GeForce GTX TITAN X as well as NVIDIA A100.

\section{Additional Experiments}
\subsection{Max-Margin Loss}
\label{app:max-margin}
We experimented with two variants of max-margin loss described in ~\citet{bhat-etal-2019-margin}, namely Syn-margin by projection (SMP) and Syn-margin by difference (SMD) on the \langpair{en}{ro} dataset. 
Using the same hyperparameters as for cosine and discrete models ($\alpha$=1, learning rate of \(10^{-4}\), and effective batch size of 65536) all max-margin models obtained scores below the best cosine model. \cref{tab:max-margin-loss} shows comparison of the models' performance when using max-margin loss and cosine loss for training CoNMT on \texttt{newstest2016} \langpair{ro}{en}.
%While further tuning might reduce these differences, we expect any changes to be less than due to the target embeddings.
\begin{table}[]
    \centering
    \begin{tabular}{l c c}
    \toprule
model & BLEU & BERT \\\midrule
cosine pre-trained (beam=1) & 29.0 & 58.5 \\
cosine pre-trained (beam=5) & 29.0 & 58.0 \\
cosine random (beam=1) & 28.0 & 58.2 \\
cosine random (beam=5) & 28.8 & 58.8 \\ \midrule
SMP pre-trained & 27.1 & 54.7 \\
SMD pre-trained & 28.5 & 57.5 \\
SMP random & 16.7 & 36.7 \\
SMD random & 26.3 & 54.3 \\
\bottomrule
    \end{tabular}
    \caption{Compariosn between cosine and max-margin loss for \texttt{newstest2016} \langpair{ro}{en}.}
    \label{tab:max-margin-loss}
\end{table}
While these results may improve with tuning, it seems unlikely for the effect to be more important than the embedding choice, and our finding that random embeddings are at least competitive with pre-trained ones holds. 
The cosine loss remains a performant, simple, and robust training objective for CoNMT with a probabilistic interpretation, making
it suitable for principled beam search, and thus we restrict the scope of our experiments to it.
%Given its performance, robustness to new architectures, simplicity, and probabilistic interpretation, we believe it’s justified to limit the scope of our current paper to studying cosine loss.

\subsection{Embeddings Dimensionality}
\label{app:embeddings-dim}
\begin{figure}
    \centering
    \includegraphics[width=.48\textwidth]{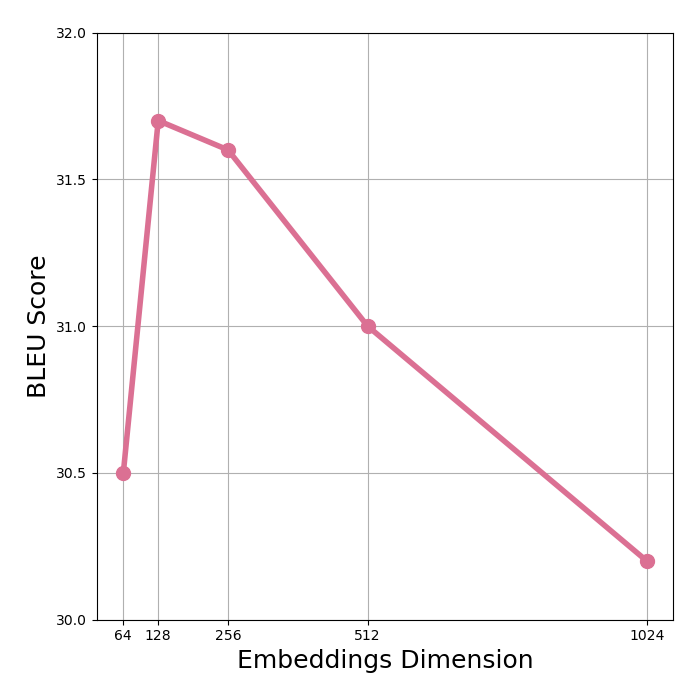}
    \caption{BLEU score of the discrete NMT models on \texttt{newstest2016} \langpair{ro}{en}.}
    \label{fig:discrete-dim}
\end{figure}

\begin{figure}[ht]
    \centering
    \includegraphics[width=\linewidth]{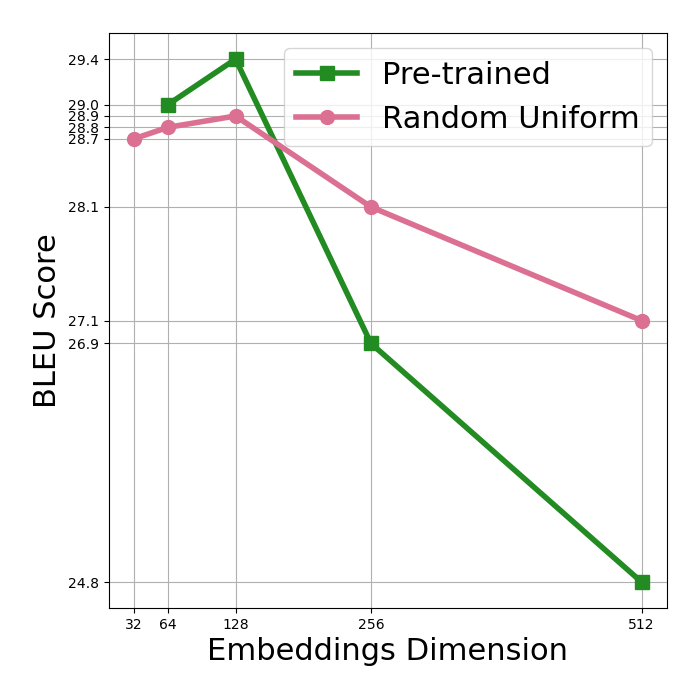}
    \caption{BLEU score on \textbf{\langpair{ro}{en}} \texttt{newstest2016} of continuous-output model with various dimensionalities of random and pre-trained target embeddings.}
    \label{fig:cont-dim}
\end{figure}

Even though it is typical to train NLP models with large embeddings dimension ($d\geq 512$), we conducted experiments on \langpair{ro}{en} and found that smaller dimensionality works better for CoNMT both with random and pre-trained target embeddings~\cref{fig:cont-dim}, and do not harm the performance of discrete model as per ~\cref{fig:discrete-dim}.

We hypothesise that better performance of lower dimensional embeddings on CoNMT is a direct consequences of the cosine distance as a distance measure. Despite its popularity, there is evidence that cosine loss is not a suitable choice for measuring the dissimilarity between high-dimensional embeddings vectors~\citep{zhou-etal-2022-problems}, and using another distance metric can potentially improve the results of the models with larger embeddings dimensionality. We leave this question for the future investigation. Since the dimensionality 128 performs the best among all tested dimensionalities, we do all our experiments with dimension equal to 128.

\subsection{Combined Embeddings}
\label{app:combined-emb}
\begin{figure}[ht!]
    \centering
    \includegraphics[width=0.9\linewidth]{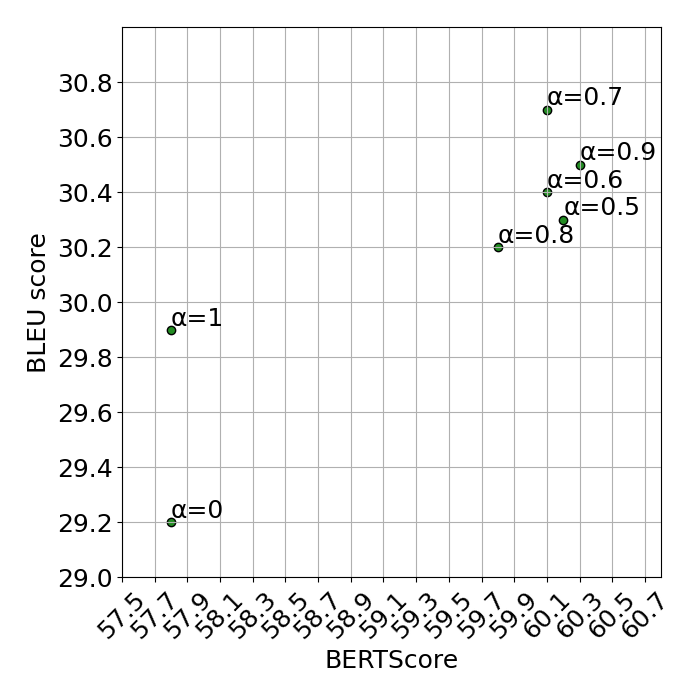}
    \caption{BLEU and BERTScores on \langpair{ro}{en} \texttt{newsdev2016}  with different values of $\alpha$.}
    \label{fig:bleu-bert-convsum}
\end{figure}

In ~\cref{tab:main results} we report performance of \texttt{combined} embeddings with $\alpha=0.9$. To study the effect of $\alpha$ on the models' performance, we conduct experiments on \langpair{ro}{en} for $\alpha \in [0.5,0.9]$. As shown in ~\cref{fig:bleu-bert-convsum}, for all cases \texttt{combined} embeddings outperform pre-trained and random ones on both metrics.

\subsection{Word Embeddings for CoNMT}
\label{app:word-level-emb}
Since the continuous-output model struggles with subwords continuation and, at the same time, performs better on rare words, we conduct experiments on the word level. Word-level model tends to suffer from out-of-vocabulary issues (\cref{tab:data-stats}), so discrete model performance drops respectively.
~\cref{tab:word-level} provides the comparison between the discrete word-level model and continuous-output model with random targets. Even though the continuous-output model struggles with subwords continuations, overall, using subwords allows us to have a stronger model both for discrete and continuous-output cases.

\begin{table}[ht]
    \centering
    \resizebox{.9\columnwidth}{!}{%
    \begin{tabular}{l c c}
    \toprule
     model & \langpair{ro}{en}&\langpair{en}{tr} \\
     \midrule
        discrete words & 28.5 & 8.9 \\
        continuous random words& 27.6 & 5.6 \\
        discrete tokens & 32.1 & 12.7\\
        continuous random tokens & 29.2 & 9.3\\
        \bottomrule
    \end{tabular}%
    }
    \caption{BLEU scores for word level and tokens level models on validation set with greedy decoding.}
    \label{tab:word-level}
\end{table}

\subsection{Subword Embeddings for CoNMT}
\label{app:subword-emb}
We rely on the unigram language model for subword segmentation~\citep{kudo-2018-subword} to train discrete and continuous-output NMT models as mentioned in Section~\cref{sec:results}. We hypothesize that it is harder for the continuous-output model to predict subwords than for the discrete model. \cref{tab:f1-spm} illustrates that the f1 macro average for the beginning of the spm tokens and continuation of the spm tokens differ a lot for discrete and continuous models. While the discrete model performs better on continuations, continuous models struggle with continuations of subwords. However, overall scores for pre-trained and random targets are the same for continuation and random embeddings performs slightly better on the beginning of the subwords.

\begin{table}[ht]
    \centering
    \resizebox{\columnwidth}{!}{%
    \begin{tabular}{l c c}
    \toprule
        \multirow{2}{*}{model}& \multicolumn{2}{c}{F1} \\ \cline{2-3}
         &  SPM start & SPM cont. \\ \midrule
       discrete  &  0.12 & 0.14 \\
       pre-trained embeddings & 0.10 & 0.09 \\
       random embeddings & 0.11 & 0.09\\
       \bottomrule
    \end{tabular}%
    }
    \caption{F1 score on \texttt{newstest2016} \langpair{ro}{en} for beginning and continuation of the SentencePiece tokens.}
    \label{tab:f1-spm}
\end{table}

% Word-level results are presented in Appendix ~\cref{app:word-level-emb}.

% \section{LSH projection to the hypercube}
% \begin{figure}[ht]
%     \centering
%     \includegraphics[width=\linewidth]{projected-cube-freq-128.png}
%     \caption{Pre-trained \langpair{ro}{en} embedddings projected onto the hypercube using LSH method. Vocabulary index is sorted by frequency, where smaller index corresponds to higher frequency.}
%     \label{fig:enter-label}
% \end{figure}

% \section{F-measure for Frequent and Rare Tokens}
% \label{app:fmeas-freq}
% \et{add fmeas plots for en-de/en-tr}
% ~\cref{fig:cosine-sim-freq} shows the dependency between token rank in the vocabulary (low rank means high frequency) and cosine similarity of the nearest and median embeddings. The trend is similar for all language pairs: the higher the rank, the higher cosine similarity, meaning that low-frequency words are clustered together more than high frequency words.

\section{Beam Search}
\label{app:beam-search}
%In our work, we use implementation of the beam search provided by \texttt{fairseq}. However, insetad of using log probabilities of the next token, we rely on the cosine similarity scores between output vector and all tokens in the vocabulary. We restrict maximum length of generated sentence to be length og the source sentence plus 200. 
Implementing beam search meaningfully for CoNMT is possible by using the following probabilistic interpretation
of the cosine loss as a Langevin:
log-likelihood with constant concentration parameter \(\kappa\): in beam search
we use this probabilistic interpretation and take
\[ \log p(y_i=t \mid \bm{y}_{<i}, \bm{x}) = -\cos(\bm{E}(t), \bm{h}) + \log C_d(1),\]
\ie, we apply the
normalizing constant of the Langevin distribution for dimension \(d\) and fixed concentration \(\kappa=1\).
We may then use the built-in \texttt{fairseq} beam search using this log-likelihood. We limit the maximum translation length to
source length plus 200.

One possible explanation why 
random embeddings perform better than pre-trained, especially for beam sizes greater than one, may be related to
disentanglement: If the continuous output prediction is “off-target” by enough to cause the nearest embedding to be wrong, provided sufficient separation between embeddings, expanding the search to more nearest neighbors can recover the solution. In contrast, for clumped pre-trained embeddings, many embeddings concentrate close to the correct one, polluting the beam.
\end{document}